\documentclass{article}
\usepackage{spconf,amsmath,graphicx}
	\usepackage{algorithm2e}
	
	\usepackage{float}
	\restylefloat{table}
	\usepackage{pseudocode}
	\usepackage{epsfig}
	\usepackage{pdfpages}
	\usepackage{graphicx}
	\usepackage{multirow}
	\usepackage{amsmath,graphicx}
	\usepackage{amssymb}
	\usepackage{subfigure}
	\usepackage{color}
	 \usepackage{sidecap}
	\usepackage{amsthm}

	\theoremstyle{definition}
	
	\theoremstyle{remark}

\newcommand{\mbf}[1]{{\mathbf #1}}

\def\bea{\begin{eqnarray}}
\def\beas{\begin{eqnarray*}}
\def\eea{\end{eqnarray}}
\def\eeas{\end{eqnarray*}}

	\def\be{\begin{equation}}
	\def\bmat{\begin{matrix}}
	\def\emat{\end{matrix}}
	\def\bea{\begin{eqnarray}}
	\def\beas{\begin{eqnarray*}}
	\def\eea{\end{eqnarray}}
	\def\eeas{\end{eqnarray*}}

	\def\bi{\begin{itemize}}
	\def\ee{\end{equation}}
	\def\ei{\end{itemize}}

	\def\z1{z^{-1}}

	\def\bmat{\begin{matrix}}
	\def\emat{\end{matrix}}

\title{Model-based free-breathing cardiac MRI reconstruction using deep learned \& STORM priors: MoDL-STORM }

\name{Sampurna Biswas, Hemant K. Aggarwal, Sunrita Poddar, and Mathews Jacob \thanks{This work is in part  supported by US  NIH  1R01EB019961-01A1  and  ONR-N000141310202 grants.}}
\address{Department of Electrical and Computer Engineering\\ The University of Iowa, IA, USA}

\begin{document}

\maketitle
	\textit{Abstract:} {We introduce a model-based reconstruction framework with deep learned (DL) and smoothness regularization on manifolds (STORM) priors to recover free breathing and ungated (FBU) cardiac MRI from highly undersampled measurements. The DL priors enable us to exploit the local correlations, while the STORM prior enables us to make use of the extensive non-local similarities that are subject dependent. We introduce a novel model-based formulation that allows the seamless integration of deep learning methods with available prior information, which current deep learning algorithms are not capable of. 
	The experimental results demonstrate the preliminary potential of this work in accelerating FBU cardiac MRI.} \\
\section{Introduction}





The acquisition of cardiac MRI data is often challenging due to the slow nature of MR acquisition. The current practice is to integrate the information from multiple cardiac cycles, while the subject is holding the breath. Unfortunately, this approach is not practical for several populations, including pediatric and obese subjects. Several self-gated strategies \cite{selfgated}, which identify the respiratory and cardiac phases to bin the data and reconstruct it, have been introduced to enable free breathing and ungated (FBU) cardiac MRI. The recently introduced, smoothness regularization on manifolds (STORM) prior in \cite{poddar2016dynamic} follows an implicit approach of combining the data from images in a data-set with similar cardiac respiratory phases. All of the current gating based methods require relatively long ($\approx$ 1 minute) acquisitions to ensure that sufficient Fourier or k-space information is available for the recovery. 

Several researchers have recently proposed convolutional neural network~(CNN) architectures for image recovery \cite{Diamond2017,unser2017,lee2017isbiJong}. A large majority of these schemes retrained existing architectures (e.g., UNET or ResNet) to recover images  from measured data. The above strategies rely on a single framework to invert the forward model (of the inverse problem) and to exploit the extensive redundancy in the images. Unfortunately, this approach cannot be used directly in our setting. Specifically, the direct recovery of the data-set using CNN priors alone is challenging due to the high acceleration (undersampling) needed ($\approx$50 fold acceleration); the use of additional k-space information from similar cardiac/respiratory phases is required to make the problem well posed. Here, high acceleration means reduced scan time which is achieved with undersampling. None of the current CNN image recovery schemes are designed to exploit such complementary prior information, especially when the prior depends heavily on cardiac and respiratory patterns of the specific subject. Another challenge is the need for large networks with many parameter to learn the complex inverse model, which requires extensive amounts of training data and significant computational power. More importantly, the dependence of the trained network on the acquisition model makes it difficult to reuse a model designed for a specific undersampling pattern to another one. This poses a challenge in the dynamic imaging setting since the sampling patterns vary from frame to frame.

 The main focus of this work is to introduce a model based framework, which merges the power of deep learning with model-based image recovery to reduce the scan time in FBU cardiac MRI. Specifically, we use a CNN based plug-and-play prior. This approach enables easy integration with the STORM regularization prior, which additionally exploits the subject dependent non-local redundancies in the data. Since we make use of the available forward model, a low-complexity network with a significantly lower number of trainable parameters is sufficient for good recovery, compared to black-box (CNN alone) image recovery strategies; this translates to faster training and requires less training data. More importantly, the network is decoupled from the specifics of the acquisition scheme and is only designed to exploit the redundancies in the image data. This allows us to use the same network to recover different frames in the data-set, each acquired using a different sampling pattern. The resulting framework can be viewed as a iterative network, where the basic building block is a combination of a data-consistency term and a CNN; unrolling the iterative network yields a network. Since it is impossible to acquire fully sampled FBU data, we validate the framework using retrospective samples of STORM  \cite{poddar2016dynamic} reconstructed data, recovered from 1000 frames (1 minute of acquisition). In this work, we aim to significantly shorten the acquisition time to 12-18 secs (200-300 frames) by additionally capitalizing on the CNN priors.
 
  The main difference between the proposed scheme and current unrolled CNN methods \cite{schlemper2017cascadeRueckert,hammernik} is the reuse of the CNN weights at each iteration; in addition to reducing the trainable parameters, the weight reuse strategy yields a structure that is consistent with the model-based framework, facilitating its easy use with other regularization terms. In addition, the use of the CNN as a plug and play prior rather than a custom designed variational model \cite{hammernik} allows us to capitalize upon the well-established software engineering frameworks such as Tensorflow and Theano. 
	\section{Proposed method} 
	We note that the STORM prior does not exploit the local similarities, such as smooth nature of the images or the similarity of an image with its immediate temporal neighbors. In contrast, the  3-D CNN architecture is ideally suited to exploit the local similarities in the dynamic data-set. To capitalize on the complementary benefits of both methods, we introduce the model based reconstruction scheme to recover the free breathing dynamic data-set $\mathbf X $ 
\begin{eqnarray}\nonumber
	\label{maineqns}
	\nonumber
	 \mathcal C(\mbf X) &=& \underbrace{\|\mathcal A(\mbf X)-\mbf B\|_2^2}_{\mbox{data consistency}} ~+~ \lambda_1 \underbrace{\|\mathcal N_{\mbf w}(\mbf X)\|^2}_{\mbox{CNN prior}}  \\&&+\qquad\lambda_2 \underbrace{{\rm tr}\left(\mathbf X^T\mathbf L\mathbf X\right)}_{\mbox{STORM prior}}. 
	\end{eqnarray}	
Here, $\mathcal A$ is the Fourier measurement operator. The second and third terms are the regularization terms, which exploit the redundancy in the images. $\mathcal N_{\rm w}$ is a learned CNN estimator of noise and alias patterns, which depends on the learned network parameters $\mathbf w$. Specifically,  $\|\mathcal N_{\mathbf w}(\mathbf x)\|^2$ is high when $\mathbf x$ is contaminated with noise and alias patterns; its use as a prior will encourage solutions that are minimally contaminated by noise and alias patterns. Since $\mathcal N_{\mathbf w}(\mathbf X)$ is an estimate of the noise and alias terms, one may obtain the \textit{denoised} estimate as
\begin{equation}
  \mathcal D_{\mathbf w} (\mathbf X) = \left(\mathcal I -\mathcal N_{\rm w}\right)(\mathbf X) = \mathbf X-\mathcal N_{\mathbf w}(\mathbf X).
\end{equation}
This reinterpretation shows that $\mathcal D_{\mathbf w}$ can be viewed as a denoising residual learning network. When $\mathcal D_{\mathbf w}$ is a denoiser, $\mathcal N_{\mathbf w}(\mathbf X)=\mathbf X - \mathcal D_{\mathbf w}(\mathbf X)$ is the residual in $\mathbf X$. With this interpretation, the CNN prior can be expressed as $\|\mathbf X - \mathcal D_{\mathbf w}(\mathbf X)\|^2$. 

The third term in \eqref{maineqns} is the STORM prior, which exploits the similarities between image frames with the same cardiac and respiratory phase; $\rm tr$ denotes the trace operator.  Here, $\mathbf L = \mathbf D-\mathbf W$ is the graph Laplacian matrix, estimated from the k-space navigators \cite{poddar2016dynamic}. $\mathbf W$ is a weight matrix, whose entries are indicative of the similarities between image frames. Specifically, $\mbf W_{(i,j)}$ is high if $\mbf x_i$ and $\mbf x_j$ have similar cardiac/respiratory phase. $\mathbf D$ is a diagonal matrix, whose diagonal entries are given by $\mathbf D_{(i,i)} = \sum_{j} \mathbf W_{(i,j)}$.

\subsection{Implementation}

\begin{figure}[t!]
     \subfigure[{Proposed model based framework }]{\includegraphics[width=\linewidth]{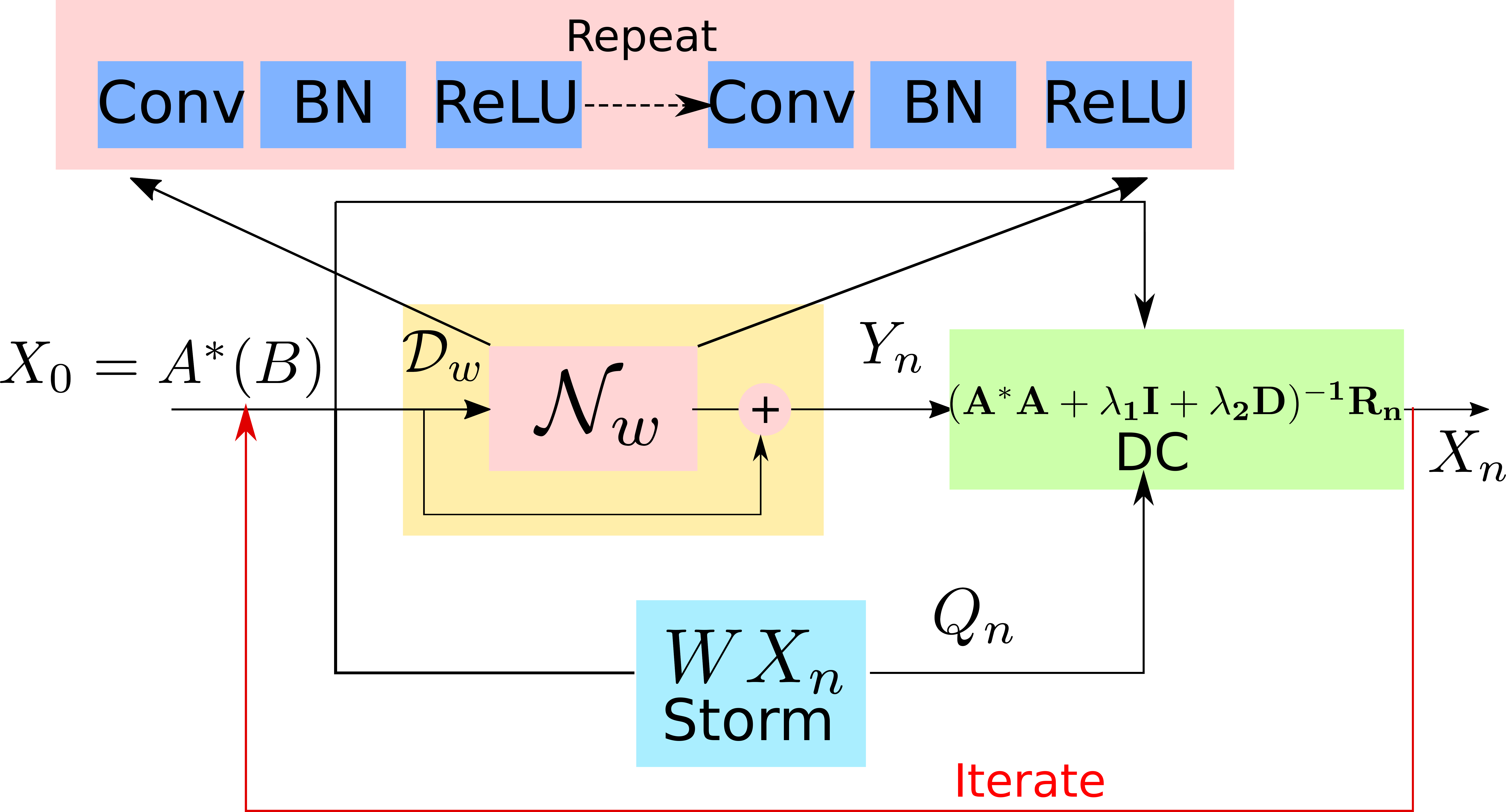}}\vspace{1em}
     \subfigure[{Unfolded network }]{\includegraphics[width=\linewidth]{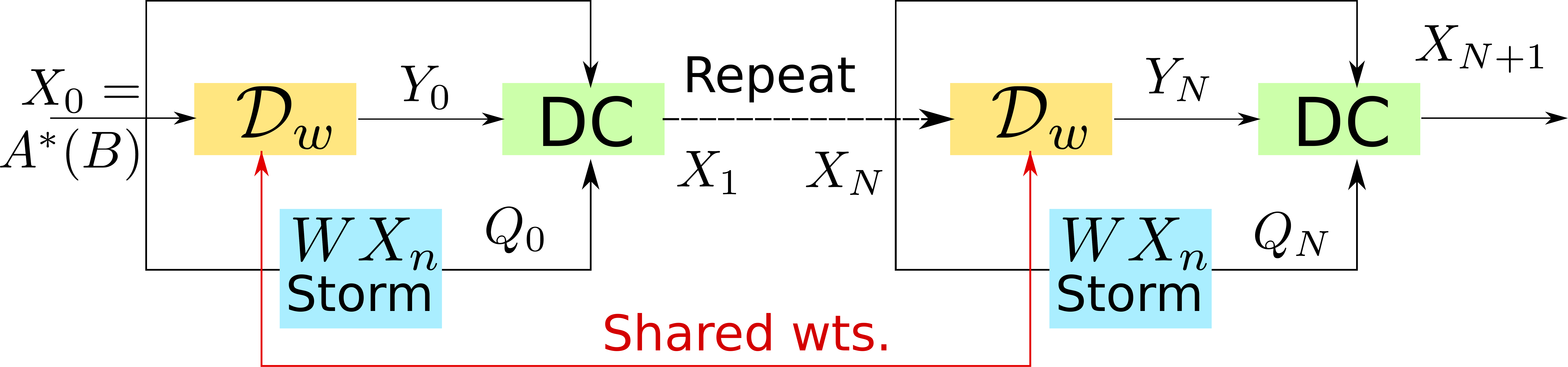}}
   \caption{Illustration of the MoDL-STORM architecture. (a) indicates the iterative structure denoted by \eqref{cnnupdate}-\eqref{dc}. Here, $\mathcal D_w$ denotes the denoising residual network, shown with the yellow box. This residual learning is performed using an out of the box CNN approach.  (b) Once the number of iterations $N$ is fixed, the network can be unrolled to obtain a non-iterative unrolled deep learning model, which relies on weight sharing.} \vspace{-0.5em}
\label{folded}
\end{figure}
We re-express the STORM prior in \eqref{maineqns} using an auxiliary variable $\mbf X=\mbf Z$:
\begin{equation}
2{\rm tr}(\mathbf X^T\mathbf L\mathbf X) = {\rm tr}(\mathbf X^T\mathbf D\mathbf X) + {\rm tr}(\mathbf Z^T\mathbf D\mathbf Z)  - 2{\rm tr}(\mathbf X^T\mathbf W\mathbf Z)
\end{equation}
This alternate expression allow us to rewrite \eqref{maineqns} using two auxiliary variables $\mbf Y = \mathcal D_{\mathbf w}(\mathbf X)$ and $\mbf Z=\mbf X$:
\begin{eqnarray*}\nonumber
	 \mathcal C &=& \|\mathcal A(\mbf X)-\mbf B\|_2^2 + \lambda_1\|\mbf X - \mbf Y\|^2+\\&&\qquad \lambda_2 \left( {\rm tr}(\mathbf X^T\mathbf D\mathbf X) + {\rm tr}(\mathbf Z^T\mathbf D\mathbf Z)  - 2{\rm tr}(\mathbf X^T\mathbf W\mathbf Z)\right)	\end{eqnarray*}	

We propose to use an update strategy, where we assume $\mbf Z$ and $\mbf Y$ to be fixed and minimize the expression with respect to $\mbf X$. The variables $\mbf Z$ and $\mbf Y$ are then updated based on the previous iterate. The minimization of $\mathcal C$ w.r.t $\mbf X$ yields
\begin{equation}
\nabla C = \mathcal A^*(\mathcal A (\mbf X)-\mbf B)+ \lambda_1 (\mbf X-\mbf Y)+ \lambda_2 \left(\mbf D\mbf X-\mbf W\mbf Z\right) =0
\end{equation}
where $\mathcal A^*$ is the adjoint of the operator $\mathcal A$. This normal equation 
which gives the update rules
\begin{eqnarray}
\label{cnnupdate}
\mbf Y_n &=&  \mathcal D_{\mbf w}(\mbf X_{n})\\\label{poddar2016dynamic}
\mbf Q_n &=&  \mbf W \mbf Z_n = \mbf W \mbf X_n\\ 
\mbf R_n &=& \left(\mathcal A^*(\mbf B) + \lambda_1\mbf Y_n + \lambda_2\mbf Q_n\right)\\
\label{dc}
\mathbf X_{n+1} &=&  (\mathcal A^*\mathcal A +\lambda_1 \mbf I+\lambda_2 \mbf D)^{-1}\mathbf R_n
\end{eqnarray}
Note that since we rely on 2-D sampling on a frame by frame basis and because $\mbf D$ is a diagonal matrix, the matrix $(\mathcal A^*\mathcal A +\lambda_1 \mbf I+\lambda_2 \mbf D)$ is separable; it can be solved analytically on a frame by frame basis in the Fourier domain; we term this operation as the data consistency (DC) step. Specifically, the update rule for the $i^{\rm th}$ frame is given by
\begin{equation}
\label{framebyframe}
\mbf x_{n+1}^{(i)} = \left[\mbf A_i^H\mbf A_i +\left(\lambda_1+\lambda_2 d_{i}\right) \mbf I\right]^{-1}  (\mbf A_i^H\mbf B + \lambda_1\mbf y_{n}^{(i)} + \lambda_2\mbf q_{n}^{(i)})
\end{equation}
where $\mbf A_i = \mbf S_i\mbf F$ and $d_{i}=\mbf D_{i,i}$. Here, $\mbf S_i$ is the sampling pattern in the $i^{\rm th}$ frame and $\mbf F$ is the 2D Fourier transform. $\mbf y_{n}^{(i)}$ is the $i^{\rm th}$ frame of $\mbf Y_n$, which is the CNN denoised version of the $n^{\rm th}$ iterate $\mbf X_n$. Likewise, $\mbf q_{n}^{(i)}$ is the $i^{\rm th}$ frame of $\mbf Q_n$, which is the STORM denoised version of $\mbf X_n$, with the exception that each column of $\mbf Q_n$ is scaled by $d_i$. This update rule shows that the framework is very similar to \cite{chan2017plug}, where the power of existing denoising algorithms are capitalized as  plug and play priors in a model based regularization framework. The algorithm alternates between signal recovery and denoising steps. 

This iterative algorithm is illustrated in Fig. \ref{folded}.(a), where we initialize the algorithm with $\mbf X_0=\mathcal A^*(\mathbf B)$. The STORM and CNN denoised signals $\mbf Q$ and $\mbf Y$ are then fed into the DC block specified by \eqref{dc}. Once the number of iterations are fixed, we can obtain the unrolled network, which is illustrated in Fig. \ref{folded}.(b). We observe that 5-8 iterations of the network is often sufficient for good recovery.

\subsection{Training phase}
We propose to train the unrolled CNN shown in Fig. \ref{folded}.b. Specifically, we present the network with exemplar data consisting of undersampled data $\mbf B$ as the input and their reconstructions as the desired output. The CNN parameters are learned to minimize the reconstruction error, in the mean square sense. Unfortunately, it is difficult to present the network with a full data-set ($\approx$ 200 frames) in the training mode. Specifically, the feature maps for each layer need to be stored for the evaluation of gradients by back-propagation, which is difficult on GPU devices with limited onboard memory. 

\begin{figure}[t!]
\centering
     \includegraphics[width=\linewidth]{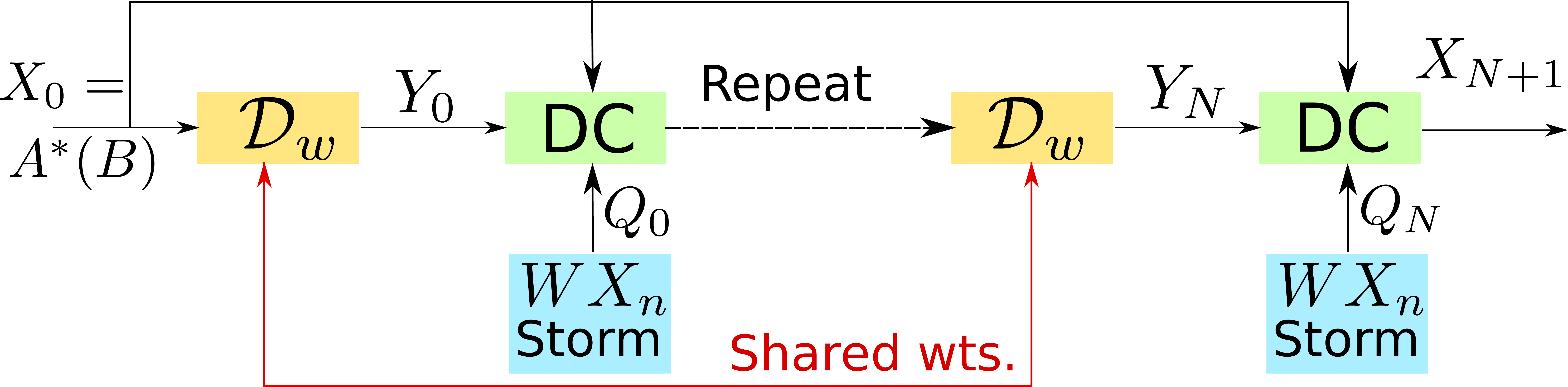}
   \caption{Modified network used in training illustrated in Algorithm \ref{pcode}. The main difference between the unrolled network in Fig.\ref{folded}.(b) is that the variables $\mbf Q_n; n=0,..,N$ are assumed to be fixed. These variables are estimated in the outer loop as shown in Algorithm \ref{pcode}.} \vspace{-1.5em}
\label{unfolded}
\end{figure}

Note that both the CNN and the data-consistency layer perform local operations. If the STORM prior was not present, one could pass smaller subsets of input and output (e.g. 20-30 frames) data in multiple batches for training. In contrast, the STORM prior uses non-local or global information from the entire data-set. Specifically, for the evaluation of the $i^{\rm th}$ frame of $\mbf Q_n$, specified by $\mbf q_{n}^{(i)}$, we require the entire data-set. We hence propose a lagged approach, where $\mbf Q_n$ is updated less frequently than the other parameters. Specifically, we use an outer loop to update $\mbf Q_n$, while the trainable parameters of the CNN $\mbf w$ and the regularization parameters $\lambda_1$ and $\lambda_2$ are updated assuming $\mbf Q_n$ to be fixed. We feed in batches of 20-30 frames of training data. The network during this training mode is illustrated in Fig. \ref{unfolded}. After one such training, we run the model on the entire data-set to  determine the $\mbf X_n; n=0,..,N-1$ for all 204 frames. The variables $\mbf Q_n$ are updated using \eqref{poddar2016dynamic}, which are then used in the next  training. 

\begin{algorithm}
\SetAlgoLined
\KwResult{$\mbf X_n$}
 Define: $\mathcal D_{\mbf w}, \mathcal A.$\ Inputs: $\mbf B,\mbf X_0=\mathcal A^*(B)$\;
 \While{$j <$ NOuter,}{
  $\mbf Q_n =  \mbf W \mbf X_n; \forall n \in [0,N]$;\\
  Train model in Fig. \ref{unfolded} with fixed $\mbf Q_n$
  $j=j+1$;
 }
 \caption{Pseudo-code for our iterative algorithm. The variables $\mbf Q_n; n=0,..,N$ are not updated while the network is trained using backpropagation. This strategy allows us to keep the memory demand of the algorithm limited, allowing us to train the network on current GPU devices. }
 \label{pcode}
\end{algorithm}

\vspace{-2em}
\section{Experimental details}
\label{ress}

The raw data data was acquired using a golden angle FLASH sequence with k-space navigators on a Siemens Aera scanner from congenital heart patients with FOV=300 mm, slice thickness=5 mm, radial views=10000, base resolution=256, TR/TE=4.68/2.1 ms, resulting in an acquisition time of $\approx 50$ s, was binned to frames with 12 lines/frame and reconstructed using the STORM algorithm using \cite{poddar2016dynamic}, where 4 channel k-space data obtained by PCA combining the data from 24 channels were used. Note that the direct acquisition and reconstruction of similar data-sets are impossible. Sub-segments of this data, each consisting of 204 frames, were used to train the proposed scheme; note that our objective is to reduce the acquisition window to around 15-20 seconds. The recovered data was retrospectively undersampled to 17 lines/frame, consisting of 4 navigator and 13 golden angle lines. The navigator lines were only used for the Laplacian estimation and not used for final reconstruction. We assumed single channel acquisition for simplicity. 

We implemented the CNN network $\mathcal D_{\mathbf w}$  in TensorFlow with a 3D kernel size of $3\times 3$, five layers, and 64 filters per layer. We considered the number of iterations of the network to be $N=8$. Each layer was a cascade of 3D convolution, batch normalization and a ReLU activation. Since the memory demands restricted the training using all 204 frames in the data-set, we split the data into batches of 17 frames each, extracted from the original data-set. Note that the sampling pattern for each  frame used in training is different; since the model based framework uses the numerical model for the acquisition scheme, it can account for this variability without requiring a very deep network with several trainable parameters.  As discussed previously and illustrated in Algorithm \ref{pcode}, we pre-compute the variables $\mbf Q_n; n=1,.., N$ and assume them to be fixed during the deep learning training procedure. We restricted the cost function to only include comparisons with the middle frames in the output to reduce boundary issues; the mean square error cost function was used to perform the optimization. We relied on ADAM optimizer with its default hyperparameters to train the network weights as well as the regularizers, $\lambda_1, \lambda_2$. We used 200 epochs of network training for each outer loop. The training time for 200 epochs is around 2.5 hours; the total time taken for  $NOuter=4$ is $\approx 11$ hours on a NVDIA P100 GPU processor.

Once the training is complete, the reconstruction follows the unfolded network in Fig. \ref{folded}.b. Specifically, the entire undersampled single channel k-space data corresponding to the 204 frame data-set is fed to the trained network. Since the network just requires eight repetitions and the basic steps involve convolutions and fast Fourier transforms, hence the reconstructions are fast. \textcolor{black}{Testing time for 204 frames with $N=8$ repetitions was 38 seconds.}

\section{Results \& Conclusion}
\begin{figure}[t!]
\centering
     \subfigure[{Ground truth}]{\includegraphics[width=.2\textwidth, height=3.5cm]{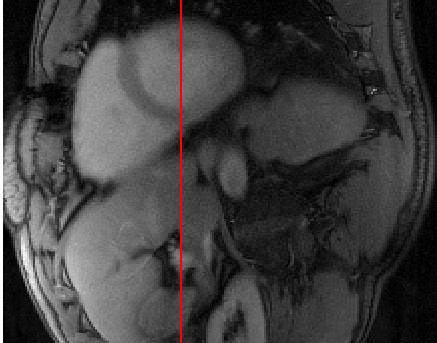}}
     \subfigure[{Gridding: SNR=6.8 dB  }]{\includegraphics[width=.2\textwidth, height=3.5cm]{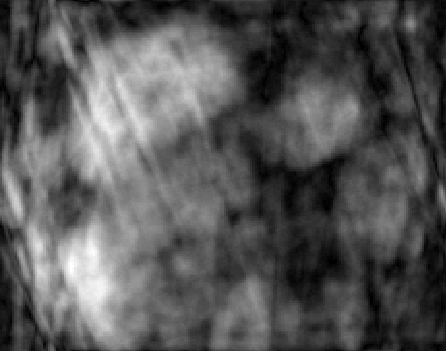}}
     \subfigure[{MoDL:SNR=13.0 dB  }]{\includegraphics[width=.2\textwidth, height=3.5cm]{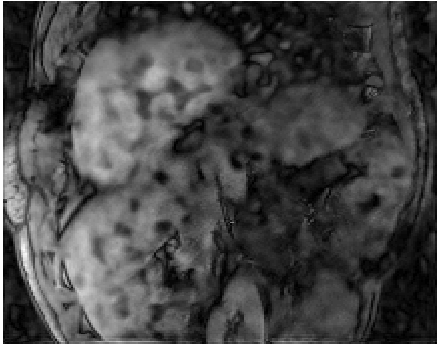}}
     \subfigure[{MoDL-STORM: SNR=23.1dB  }]{\includegraphics[width=.2\textwidth, height=3.5cm]{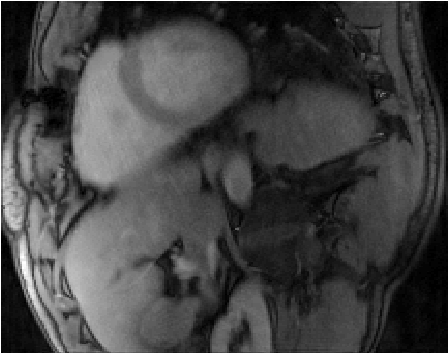}}
      \subfigure[{Ground truth}]{\includegraphics[width=.2\textwidth]{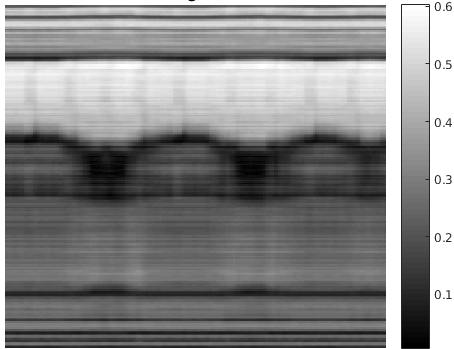}}
     \subfigure[{MoDL-STORM}]{\includegraphics[width=.2\textwidth]{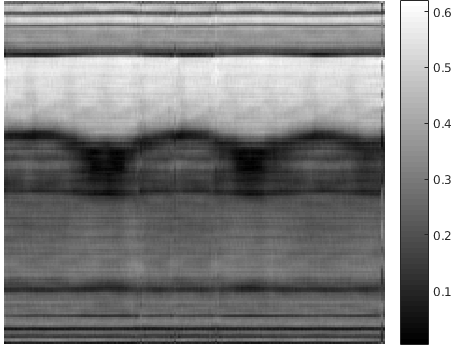}}
\caption{Reconstruction results at 16 fold acceleration. The ground truth data, recovered from 50 second (1000 frames) acquisition using STORM is shown in (a). The data from a subset of 204 frames is retrospectively downsampled assuming single channel acquisition and tested using different algorithm.
The zero-filled IFFT reconstruction is shown in (b). A reconstructed frame using a network similar to ours without STORM priors (only CNN)  is shown in (c). Note that the exploitation of the local similarities enabled by CNN alone is insufficient to provide good reconstructions at 16 fold acceleration. In contrast, the combination of deep learning and STORM priors yield reconstructions that are comparable in quality to the ground truth. The temporal profiles of the ground truth and MoDL-STORM reconstructions are shown in (e) and (f), respectively}
\label{rec16x}
\end{figure}

Fig \ref{rec16x}  shows the reconstruction results with corresponding to 17 lines/frame) using single channel data.  This data is used as the ground truth for the reconstruction. A similar STORM reconstructed data-set from another patient was used for training. These comparisons show that the combination of deep learning and STORM priors yield reconstructions that are comparable in quality to the ground truth. 

We introduced model based dynamic MR reconstruction for free breathing and ungated cardiac MRI. The proposed framework enables the seamless integration of deep learned architectures with other regularization terms. It additionally exploits the prior information that is subject dependent (e.g. due to respiratory variations and cardiac rate). The preliminary study in this paper shows that the proposed framework enables us to significantly reduce the acquisition time in free breathing approaches. Our future work will focus on demonstrating the power of the framework on prospectively acquired data. We also plan to capitalize on the multichannel data, which is not exploited in this work for simplicity.

\bibliographystyle{IEEEtran}
\bibliography{refs}

	\end{document}